\documentclass[letterpaper]{article} 
\usepackage{aaai24}  
\usepackage{times}  
\usepackage{helvet}  
\usepackage{courier}  
\usepackage[hyphens]{url}  
\usepackage{graphicx} 
\usepackage{amsfonts}
\usepackage{amsmath}
\usepackage{natbib}  
\usepackage{caption} 
\usepackage{algorithm}
\usepackage{algorithmic}
\usepackage{subfigure}
\usepackage{newfloat}
\usepackage{listings}
\DeclareCaptionStyle{ruled}{labelfont=normalfont,labelsep=colon,strut=off} 
\floatstyle{ruled}
\newfloat{listing}{tb}{lst}{}
\floatname{listing}{Listing}
\title{Point Deformable Network with Enhanced Normal Embedding \\
for Point Cloud Analysis}
\author{
    Xingyilang Yin\textsuperscript{\rm 1},
    Xi Yang\textsuperscript{\rm 1}\thanks{Corresponding author.},
    Liangchen Liu\textsuperscript{\rm 1},
    Nannan Wang\textsuperscript{\rm 1},
    Xinbo Gao\textsuperscript{\rm 2}
}

\affiliations{
    \textsuperscript{\rm 1}Xidian University\\

    \textsuperscript{\rm 2} Chongqing University of Posts and Telecommunications\\
    yxyl@stu.xidian.edu.cn, \{yangx, nnwang\}@xidian.edu.cn, lcliu79xidian@gmail.com, gaoxb@cqupt.edu.cn
}

\usepackage{bibentry}

\begin{document}

\maketitle

\begin{abstract}
Recently MLP-based methods have shown strong performance in point cloud analysis. Simple MLP architectures are able to learn geometric features in local point groups yet fail to model long-range dependencies directly. In this paper, we propose Point Deformable Network (PDNet), a concise MLP-based network that can capture long-range relations with strong representation ability. Specifically, we put forward Point Deformable Aggregation Module (PDAM) to improve representation capability in both long-range dependency and adaptive aggregation among points. For each query point, PDAM aggregates information from deformable reference points rather than points in limited local areas. The deformable reference points are generated data-dependent, and we initialize them according to the input point positions. Additional offsets and modulation scalars are learned on the whole point features, which shift the deformable reference points to the regions of interest. We also suggest estimating the normal vector for point clouds and applying Enhanced Normal Embedding (ENE) to the geometric extractors to improve the representation ability of single-point. Extensive experiments and ablation studies on various benchmarks demonstrate the effectiveness and superiority of our PDNet.
\end{abstract}

\section{Introduction}
\begin{figure}
	\centering
	\subfigure[kNN]{
		\begin{minipage}[b]{0.14\textwidth}
			\includegraphics[width=1\textwidth]{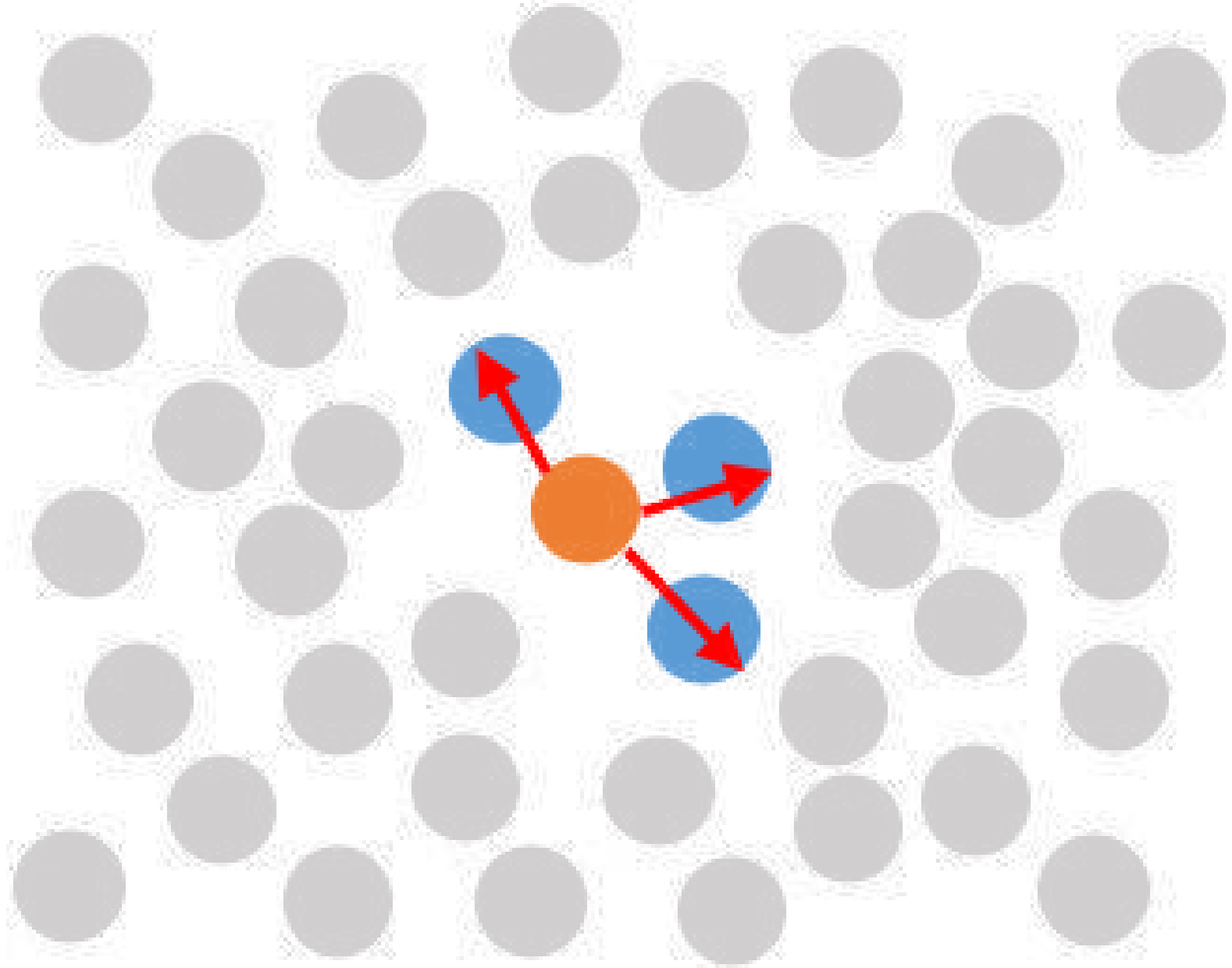}
		\end{minipage}
		\label{fig:1a}
	}
    	\subfigure[ball query]{
    		\begin{minipage}[b]{0.14\textwidth}
   		 	\includegraphics[width=1\textwidth]{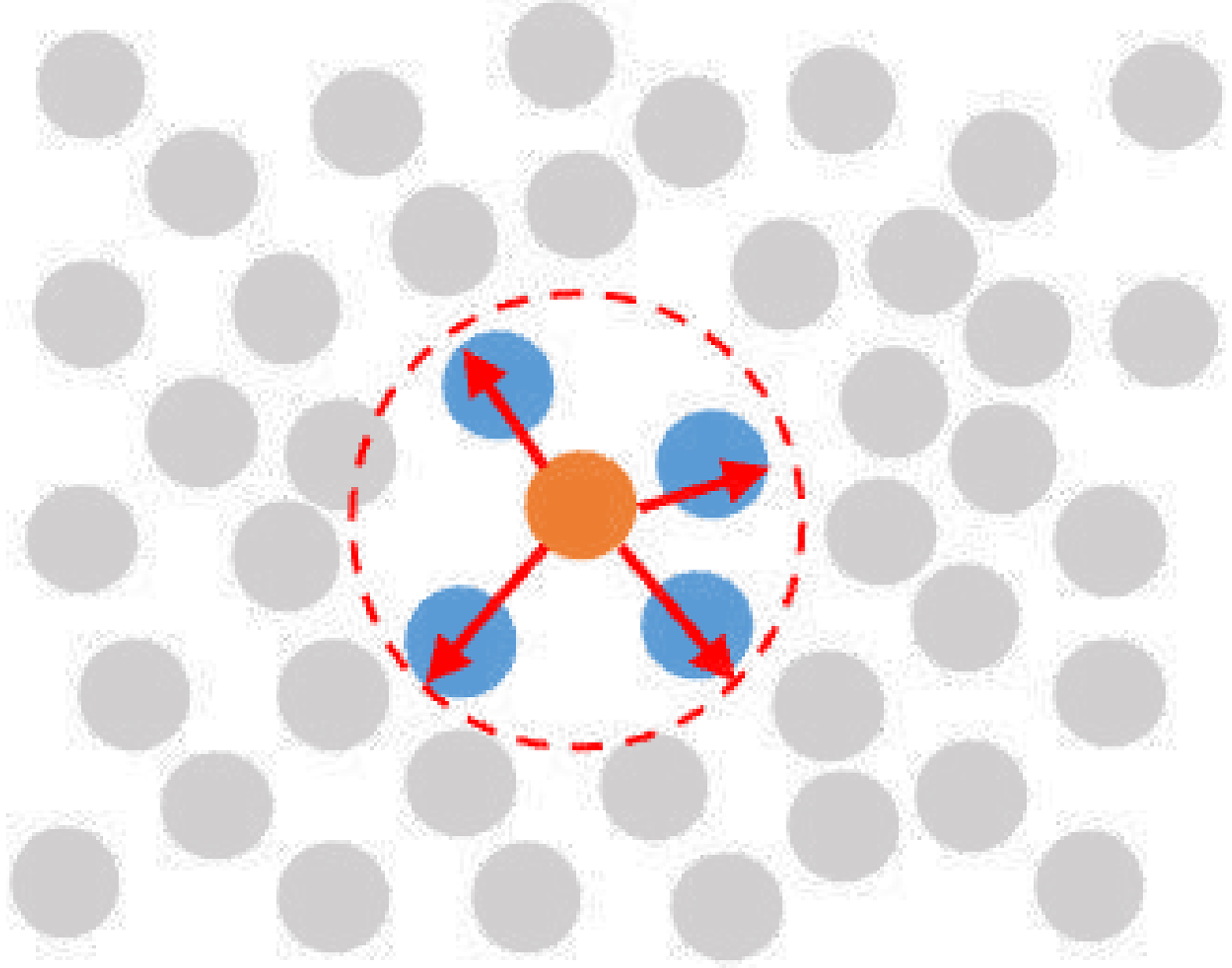}
    		\end{minipage}
		\label{fig:1b}
        }
            \subfigure[deformable]{
    		\begin{minipage}[b]{0.14\textwidth}
   		 	\includegraphics[width=1\textwidth]{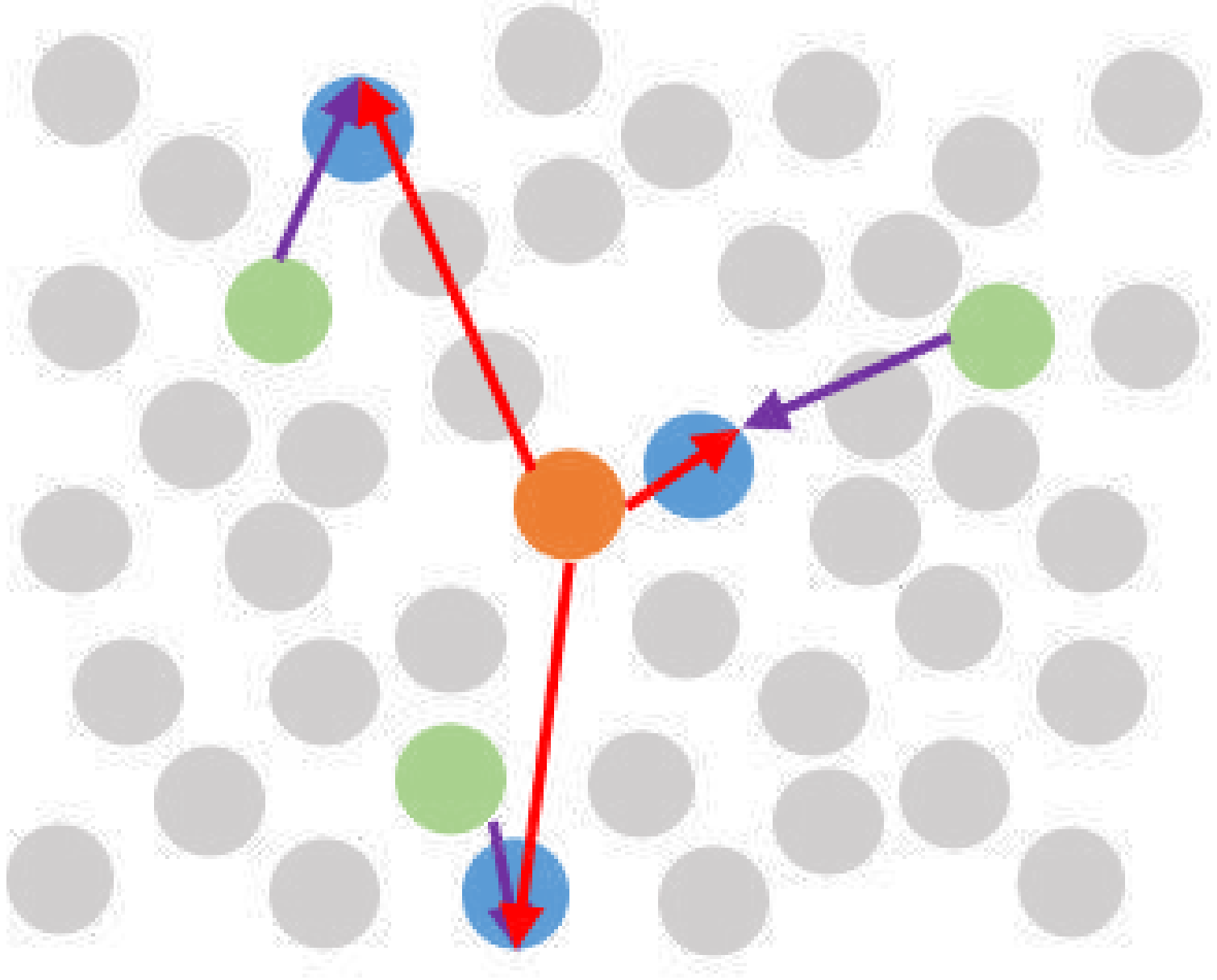}
    		\end{minipage}
		\label{fig:1c}
        }
	\caption{Comparison of grouping methods. (a) kNN: query point (orange point) and its $k$-nearest ($k$=3) neighbors (blue points). (b) ball query: query point and its neighbors in local region (red circle). (c) query point and its deformable reference points (blue points) located in non-local regions. The green points and purple arrows represent the initial reference points generated on all point positions and their learnable offsets conditioned on the whole point features, respectively.}
	\label{fig:point_deform_illustration}
\end{figure}

Point cloud analysis receives great interest due to numerous 3D data acquisition devices applied in various areas, such as autonomous driving and robotics. Unlike images that have regular 2D grids, point clouds are inherently sparse, unordered, and unstructured data. Thus directly processing point clouds is challenging. Recently, MLP-based methods have obtained significant performance with simple components. PointNeXt \cite{qian2022pointnext} revisits the classical PointNet++ \cite{qi2017pointnet++} and improves it with modern training and scaling strategies. PointNeXt makes enormous improvements compared to the PointNet++ and even outperforms dedicated designed convolution-based \cite{xu2021paconv}, graph-based \cite{zhou2021adaptive}, and powerful point transformers methods \cite{zhao2021point, lai2022stratified, wu2022point}. This reveals that concise MLP modules can already describe the local geometric properties of point clouds. The following PointMetaBase \cite{lin2023meta} further modifies PointNeXt with explicit position encoding and MLP before grouping operation. Although the existing MLP-based approaches show strong generalization ability in various tasks, they ignore modeling long-range dependencies. As illustrated in Figure \ref{fig:1a} and \ref{fig:1b}, the previous MLP-based methods \cite{qi2017pointnet++, ma2022rethinking, qian2022pointnext, lin2023meta} only focus on aggregating information in local point groups constructed by kNN or ball query, which fails to learn features from a long distance. However, capturing long-range relations has been demonstrated to be crucial in understanding global shape context \cite{wang2018non, lai2022stratified}.

To solve the aforementioned problem, we need to explore aggregating information from distant regions for each query point. In the literature on processing images, learning deformable convolution filters has been shown effective in various challenging vision tasks due to adaptive spatial aggregation in long-range and more informative regions \cite{dai2017deformable, zhu2019deformable, zhu2020deformable, wang2023internimage}. This motivates us to design the deformable mechanism for point clouds. In contrast to the images that have structured 2D grids, point clouds are sparse and unstructured data. Thus naive implementation of deformable mechanisms suited for images can not directly apply to point clouds. To alleviate it, KPConv \cite{thomas2019kpconv} first introduces a deformable mechanism to point clouds. It adopts pseudo-grid convolution through predefined kernel points based on local point positions with weight matrices learned by local point features. The deformable mechanism is further applied to learn offsets constrained in local areas, which helps refine the kernel points. 

However, KPConv still focuses on developing sophisticated modules to extract local structures and long-range dependency is not considered. To this end, we propose a simple and effective MLP-based network named Point Deformable Network (PDNet). Specifically, we first put forward Point Deformable Aggregation Module (PDAM), which achieves both long-range dependency and adaptive spatial aggregation that suits point clouds at the same time. As shown in Figure \ref{fig:1c}, different from extractors that aggregate information in fixed local regions, our PDAM aggregates information from deformable reference points for each query point. The initial reference points are generated on the positions of all points, and the additional offsets and modulation scalars are learned on the whole point features. Thus the reference points are shifted to the relevant regions and bring more informative geometric features for aggregation, which strengthens the representation ability among points. Further, we suggest applying the least square fitting to estimate the normal vector of a point cloud and using Enhanced Normal Embedding (ENE) to improve the representation ability of single-point. Extensive experiments on various challenging benchmarks demonstrate the effectiveness of our methods. Our PDNet outperforms other competitive MLP-based models and achieves state-of-the-art.

\section{Related Work}
\paragraph{Point-based networks on point clouds.} In contrast to the project methods that project point clouds to multi-view images \cite{su2015multi, goyal2021revisiting} or structured 3D voxel \cite{wu20153d, maturana2015voxnet}, point-based methods process unstructured point clouds directly. PointNet \cite{qi2017pointnet}, the pioneering point-based network, proposes to model the permutation invariance of point clouds by using shared MLPs to encode pointwise features and aggregating them by symmetric functions like max-pooling. To better capture local geometric structures, PointNet++ \cite{qi2017pointnet++} proposes a hierarchical structure by gradually downsampling with farthest point sampling and aggregating features from neighbor points with kNN or ball query method. Currently, most point-based methods focus on the design of local geometric extractors. Convolution-based approaches \cite{li2018pointcnn, liu2019relation, thomas2019kpconv, xu2021paconv} propose several invariant and dynamic convolution kernels to aggregate point features. Graph convolution-based methods \cite{wang2018local, wang2019dynamic, zhou2021adaptive} treat points and their relations as vertices and edges of a graph, respectively. Point features can then be extracted by applying graph convolution on the graph. Point Transformers \cite{zhao2021point, guo2021pct, lai2022stratified, wu2022point} capture local and global information through self-attention. Recently, MLP-based approaches \cite{ ma2022rethinking, ran2022surface, tang2022warpinggan, qian2022pointnext, zhang2023parameter, lin2023meta} obtain competitive results with simple network architectures. PointMLP proposes a geometric affine module to enhance the residual MLPs network. PointNeXt follows the design philosophy of PointNet++ and integrates with improved training and scaling strategies. PointMetaBase revisits the existing methods and proposes a meta-architecture for point cloud analysis. Although these MLP-based networks show high performance in learning local geometry, the exploration of long-range dependency is omitted. Our PDNet is an MLP-based network that enjoys both long-range dependency and adaptive position aggregation inspired by deformable mechanisms.
\paragraph{Deformable networks on images.} The deformable mechanism is first presented by Deformable convolutional network (DCN) \cite{dai2017deformable} to enhance the capability of convolution with additional offsets and adaptive spatial aggregation conditioned on input data. DCNv2 \cite{zhu2019deformable} improves its ability by introducing a modulation mechanism. The deformable mechanism has also been applied to ViTs \cite{zhu2020deformable, yue2021vision, xia2022vision}, which shows powerful capability in refining visual tokens. Recently, InternImage \cite{wang2023internimage} proposes large-scale ViT architecture with DCNv3, which gains both benefits in long-range dependency and adaptive spatial aggregation and outperforms related work. However, deformable mechanisms designed for images do not fit unstructured point clouds. This work aims to develop a deformable mechanism for point clouds to aggregate point features from relevant areas through learned initial positions and offsets conditioned on the input points.

\section{Methods}
In this section, we first shortly describe the background of MLP-based approaches. Second, we propose Point Deformable Aggregation Module to achieve both long-range dependency and adaptive spatial aggregation in a data-dependent way. Third, we introduce the least square fitting to estimate the point normal vector and suggest applying additional normal embedding to strengthen the representation ability of the network. Finally, we present the overall architectures of PDNet for classification and segmentation tasks.

\subsection{Preliminary} 
\label{sec:3_1}
In this subsection, we briefly revisit some point MLP-based approaches such as PointNet++ \cite{qi2017pointnet++}, PointNeXt \cite{qian2022pointnext}, and PointMetaBase \cite{lin2023meta}.

\paragraph{PointNet++} captures local geometric features through the set abstraction (SA) module. SA module consists of subsample layer to select the input points and neighborhood aggregation module to extract local patterns. The neighborhood aggregation is formulated as:
\begin{equation}
    f_{i}^{l+1} = \mathcal{A}(\{\mathcal{M}([f_{j}^{l}, p_{j}^{l}-p_{i}^{l}]), \forall j \in \mathcal{N}_i\}),
    \label{eq:1}
\end{equation}
where $\mathcal{N}_i$ is the index set of neighbors of point $i$. $p_{i}^{l}$, 
$p_{j}^{l}$, $f_{j}^{l}$ are the point coordinates selected through farthest point sampling, the coordinates, and the features of neighbor $j$ in the stage $l$ of the network, respectively. $\mathcal{M}$ represents the shared MLPs that encode the concatenation of point features of neighbor $j$ and the relative coordinates $p_{j}^{l}-p_{i}^{l}$. $\mathcal{A}$ is the symmetric aggregation function such as max-pooling.

\paragraph{PointNeXt} further appends Inverted residual MLP (InvResMLP) block after SA module to enhance point features:
\begin{equation}
    f_{i}^{l+1} = \mathcal{M}_{2}(\mathcal{A}(\{\mathcal{M}_{1}([f_{j}^{l}, p_{j}^{l}-p_{i}^{l}]), \forall j \in \mathcal{N}_i\})) + f_{i}^{l},
    \label{eq:2}
\end{equation}
where PointNeXt uses one layer MLP $\mathcal{M}_{1}$ for neighbor feature aggregation and 2-layer MLP $\mathcal{M}_{2}$ for point feature update. $f_{i}^{l}$ is the input point features in stage $l$.

\paragraph{PointMetaBase} slightly modifies InvResMLP and applies position encoding $\delta$ for relative coordinates $p_{j}^{l}-p_{i}^{l}$:
\begin{equation}
    {{f^{l}_{i}}}^{'} = \mathcal{M}_{3}(f_{i}^{l}),  {{f^{l}_{j}}}^{'} = Group({{f^{l}_{i}}}^{'}, p_{i}^{l}),
    \label{eq:3}
\end{equation}
\begin{equation}
    f_{i}^{l+1} = \mathcal{M}_{2}(\mathcal{A}(\{{{f^{l}_{j}}}^{'} + \delta(p_{j}^{l}-p_{i}^{l}), \forall j \in \mathcal{N}_i\})) + f_{i}^{l}.
    \label{eq:5}
\end{equation}
Notice that PointNeXt uses the mapping function $\mathcal{M}_{1}$ (eg: MLP) after the grouping layer while PointMetaBase adopts $\mathcal{M}_{3}$ before the grouping operation to reduce computation. ${{f^{l}_{i}}}^{'}$ and ${{f^{l}_{j}}}^{'}$ are the updated point features of point $i$ and its neighbor $j$, respectively.

\begin{figure}[t]
\centering
\includegraphics[width=0.95\columnwidth]{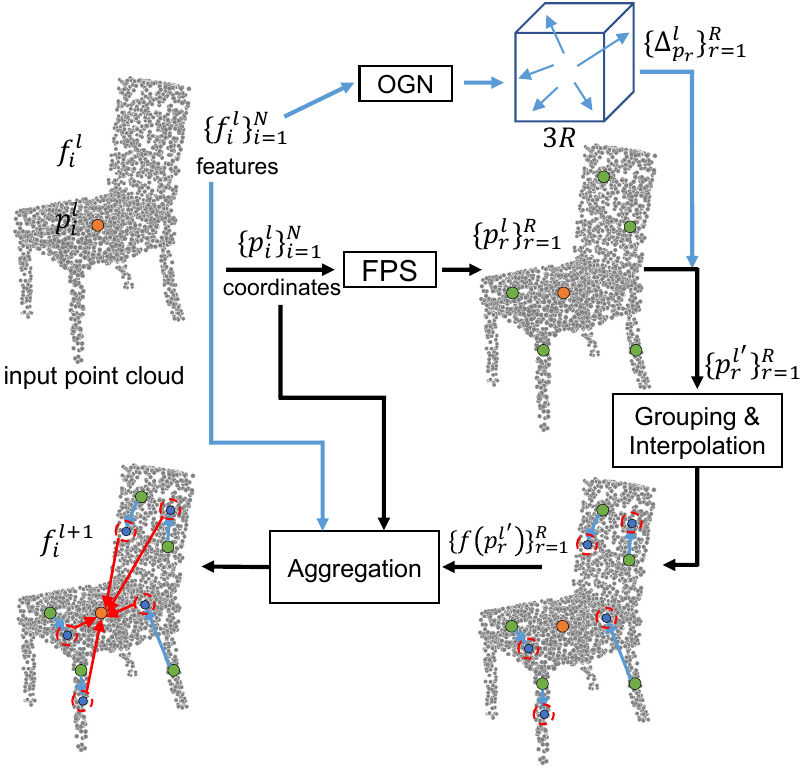} 
\caption{Illustration of Point Deformable Aggregation Module. Given input $N$ points with coordinates $\{p_i^l\}_{i=1}^N$ and features $\{f_i^l\}_{i=1}^N$, FPS generates the initial reference points (green points) based on $\{p_i^l\}_{i=1}^N$, and OGN determines the $3R$ offsets (blue arrows) learned on $\{f_i^l\}_{i=1}^N$. Then, features of adaptive position $f({{p^{l}_{r}}}^{'})$ are computed through grouping and interpolation. Finally, for point $i$, point position $p_i^l$, point feature $f_i^l$, positions $\{{{p^{l}_{r}}}^{'}\}_{r=1}^{R}$ (blue points) and features $\{f({{p^{l}_{r}}}^{'})\}_{r=1}^{R}$ of deformable reference points participate together in forming updated point features $f_{i}^{l+1}$.}
\label{fig:point_deform_illustration3d}
\end{figure}

\begin{figure*}[t]
\centering
\includegraphics[width=1.98\columnwidth]{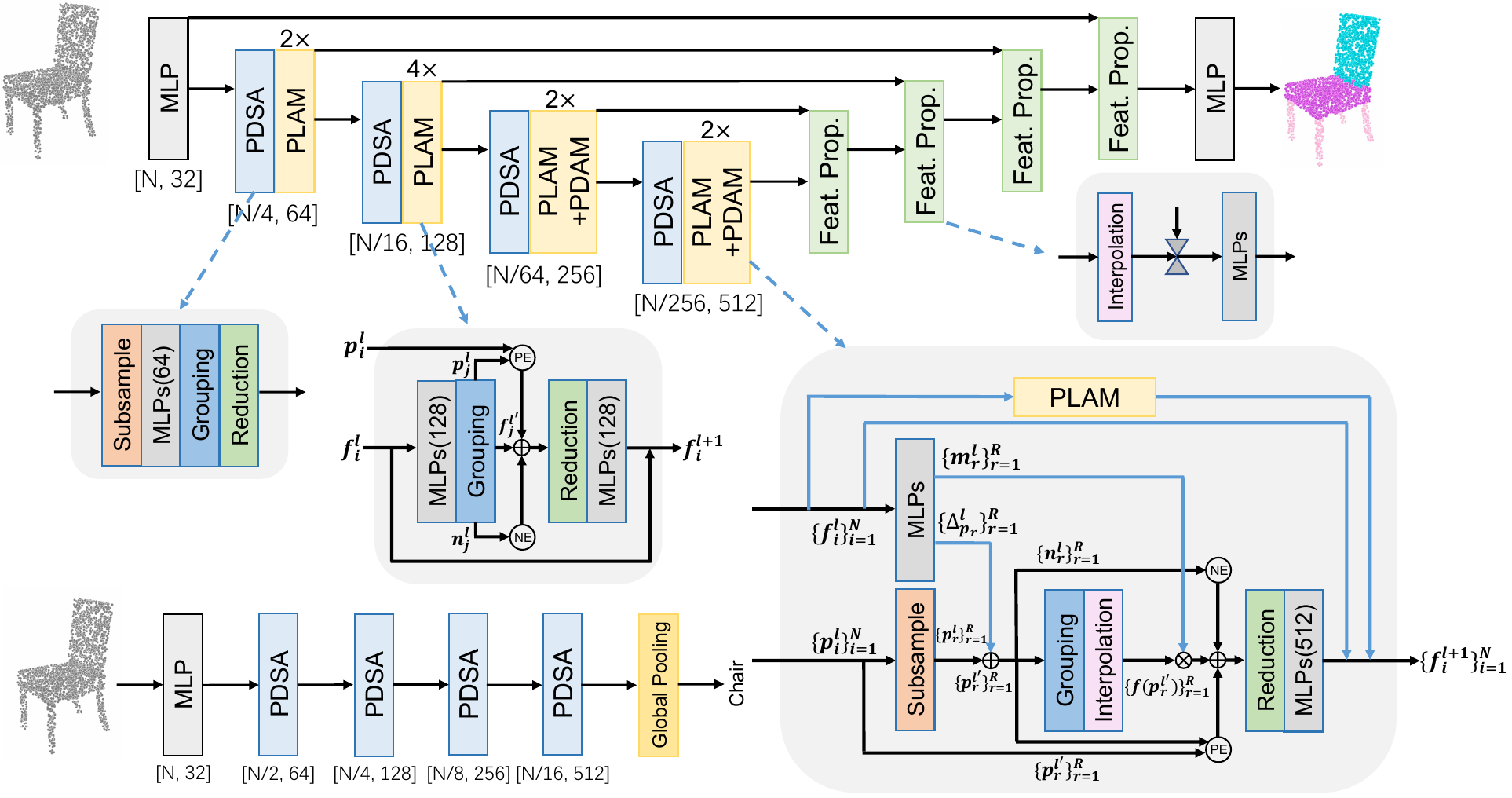} 
\caption{Illustration of Point Deformable Network (PDNet) and macro-design of PDNet-L. For classification (bottom left), we use consecutively PDSA block, which incorporate Set Abstraction module \cite{qi2017pointnet++} with position encoding and normal embedding. For segmentation (top), we adopt a U-net style architecture with Feature Propagation \cite{qi2017pointnet++} as decoder and PDSA, PLAM, and PDAM as encoder. }
\label{fig:point_deformable_network}
\end{figure*}

\subsection{Point Deformable Aggregation Module}
As discussed in Section 3.1, previous MLP-based approaches capture geometric features through local point groups. They aggregate features in local areas such as the fixed numbers of local neighbor points or points in a small radius, which fail to directly model long-range dependency. To solve it, we propose Point Deformable Aggregation Module (PDAM), which captures long-range relations and achieves adaptive spatial aggregation at the same time in a data-dependent way. Given an input image $x \in \mathbb{R}^{C\times H\times W}$, deformable mechanism proposed in DCNv2 \cite{zhu2019deformable} can be described as:
\begin{equation}
    y(p) = \sum_{k=1}^{K}w_km_kx(p+p_k+ \Delta p_k),
    \label{eq:6}
\end{equation}
where $y(p)$ and $x(p)$ denote the output feature maps and input feature maps at location $p$. $K$ represents the number of sampling locations, $w_k$ and $p_k$ are weight projection and predefined offset for the $k$-th location, respectively. For example, for a convolution with 3 $\times$ 3 kernel and dilation 1, $K$ = 9, $p_k \in \{(-1, -1), (-1, 0), ..., (1, 1)\}$. $\Delta p_k$ is the learnable offset of $2K$ channels condition on the input feature $x$. $m_k$ is the modulation scalar of $K$ channels obtained through a convolutional layer and sigmoid activation over the same input. However, applying the deformable mechanism in images to points is a non-trivial problem. In contrast to the images that have structured 2D grids, point clouds are kind of unstructured data that are unevenly distributed in space. Directly using the predefined grid sampling like $p_k$ does not suit point clouds.

Inspired by the deformable mechanism presented by \cite{dai2017deformable, zhu2019deformable}, we propose the PDAM to offer point clouds adaptive position aggregation and long-range relations via deformable reference points, as illustrated in Figure \ref{fig:point_deform_illustration3d}. Specifically, given the input point cloud with $N$ points at $l$-th stage as $\{p_i^l, f_i^l\}_{i=1}^{N}$, where $p_i^l \in \mathbb{R}^{1 \times 3}$ and $f_i^l \in \mathbb{R}^{1 \times C}$ are the coordinate and feature of point $i$, respectively. For every point $i$, we first initialize $R$ points as references through farthest point sampling (FPS) based on all the point positions, which solves the defect of irregular point cloud and leads the initial reference points $\{p_r^l\}_{r=1}^{R}$ to be uniformly distributed. This process can be described as:
\begin{equation}
    \{p_r^l\}_{r=1}^{R} = FPS(\{p_i^l\}_{i=1}^{N}).
    \label{eq:7}
\end{equation}

Different from KPConv \cite{thomas2019kpconv} that computes and refines kernel points within a local sphere, our initial reference points are based on the positions of all points, which enjoy larger receptive field. Then, to obtain the offset for each reference point, we feed the point features $\{f_i^l\}_{i=1}^N, f_i^l \in \mathbb{R}^{1 \times C}$ of all points to the offset generation network (OGN) to output the offsets $\{\Delta p_r^l\}_{r=1}^{R}, \Delta p_r^l \in \mathbb{R}^{1 \times 3}$ as the following:
\begin{equation}
            \{\Delta p_r^l\}_{r=1}^{R} = OGN(\{f_i^l\}_{i=1}^{N}),
    \label{eq:8}
\end{equation}
where OGN is implemented as two linear layers with learnable weight matrics $W_1 \in \mathbb{R}^{N \times N}$ and $W_2 \in \mathbb{R}^{ N \times 3R}$ to get 3$R$ offsets, hence the 3$R$ offsets will shift reference points to any reasonable position. It will bring the query point more relevant information in global contexts learned on the whole point features. However, since point clouds are discrete data points, there may not exist point on positions $\{{{p^{l}_{r}}}^{'}\}_{r=1}^R, {{p^{l}_{r}}}^{'} = p_r^l + \Delta p_r^l$. To alleviate it, we adopt inverse distance weighted average based on $K$ nearest neighbors (KNN) and interpolate features nearby position ${{p^{l}_{r}}}^{'}$ to get the features of deformable reference point in the local region of position ${{p^{l}_{r}}}^{'}$ as follows:
\begin{equation}
            \{p_k^l\}_{k=1}^K = KNN({{p^{l}_{r}}}^{'}),
    \label{eq:9}
\end{equation}

\begin{equation}
            f^{(j)}({{p^{l}_{r}}}^{'}) = \frac{\sum_{k=1}^K w_k({{p^{l}_{r}}}^{'})f_k^{(j)}}{\sum_{k=1}^K w_k({{p^{l}_{r}}}^{'})}, j = 1, ..., C,
    \label{eq:10}
\end{equation}
where $\{p_k^l\}_{k=1}^K$ are the $K$ nearest neighbor points of position ${{p^{l}_{r}}}^{'}$.  $w_k({{p^{l}_{r}}}^{'}) = \frac{1}{d({{p^{l}_{r}}}^{'}, p_k^l)}$ is the weighted parameter, and $d(\cdot, \cdot)$ computes the distance between two points. Thus the deformable reference points have the information $f({{p^{l}_{r}}}^{'})$ in relevant regions, which will be involved in the aggregation with the query point $i$.

Further, we use two linear layers with a sigmoid layer to obtain $R$ channels of the modulation scalars $\{\Delta m_r^l\}_{r=1}^R$ for deformable reference point features. Finally, the aggregation procedure for the query point can be defined as:
\begin{equation}
            f_{i}^{l+1} = \mathcal{M}(\mathcal{A}(\{\Delta m_r^l f({{p^{l}_{r}}}^{'}) + \delta({{p^{l}_{r}}}^{'}-p_{i}^{l})\}_{r=1}^R )) + f_{i}^{l}, 
    \label{eq:11}
\end{equation}
where $\delta({{p^{l}_{r}}}^{'}-p_{i}^{l})$ is relative position embedding of point $i$ and its deformable reference points ${{p^{l}_{r}}}^{'}$. $\mathcal{A}$ is the symmetric aggregation function (max-pooling), $\mathcal{M}$ is a mapping function such as MLP.

\subsection{Enhanced Normal Embedding}
PDAM aggregates long-range contexts from regions of interest for each query point, strengthening the representation capability among points. In this subsection, we further propose Enhanced Normal Embedding to improve the representation ability of each point itself. Normal features provide geometric information about point clouds. Using additional point normals rather than only consuming point coordinates in the network has been proven effective in various works \cite{qi2017pointnet, qi2017pointnet++, li2018so, wu2019pointconv}. However, it does not work if no point normals exist in the dataset. Inspired by \cite{mitra2003estimating}, we adopt the least square fitting to estimate the normal vector of a point cloud. Considering point $i$ and its $k-1$ nearest neighbor points $\mathcal{N}_i$, the covariance matrix $M$ can be computed as:
\begin{equation}
            M = \frac{1}{k}\sum_{i=1}^{k}(p_i-\bar{p})(p_i-\bar{p})^T, 
    \label{eq:12}
\end{equation}
where $p_i \in \mathbb{R}^{1\times3}$ is the coordinate of point $i$ in the $l$-th stage, $\bar{p} = \frac{1}{k}\sum_{i=1}^{k}p_i$ denotes the centroid of point $i$ and its neighbors. Thus $M$ is $3\times3$ symmetric positive semidefinite matrix. The normal to the local least square plane for point $i$ can be estimated as the eigenvector corresponding to the minimum eigenvalue of $M$ \cite{mitra2003estimating}. We utilize singular value decomposition to obtain the normal feature $n_i$ for each point $i$.

Like widely used position encoding in attention-based and MLP-based methods \cite{lai2022stratified, wu2022point, lin2023meta} to learn complex point cloud positional relations among point groups or all points, we propose to strengthen the point normal via Enhanced Normal Embedding (ENE). According to \cite{yang2020cn, ran2022surface}, point position and point normal are features with different distributions, which can be decoupled along channel dimension and fused through summation after embedding. In this paper, we implement ENE with 2-layer MLP.


\subsection{Point Deformable Network Architectures}
As illustrated in Figure \ref{fig:point_deformable_network}, we propose Point Deformable Network (PDNet), shared the similar hierarchical structure as \cite{lin2023meta, qian2022pointnext}, and incorporated with Point Deformable Aggregation Module (PDAM) and Enhanced Normal Embedding (ENE). For the segmentation task, we use a U-net architecture, which contains an encoder and a decoder. For the classification task, we only use an encoder. The decoder comprises widely used Feature Propagation layers \cite{qi2017pointnet++, qian2022pointnext, lin2023meta} to gradually upsample features via interpolation. Incorporating position embedding and ENE, we tweak the Set Abstraction module \cite{qi2017pointnet++, qian2022pointnext} as the reduction block, termed Point Deformable Set Abstraction (PDSA). The encoder is composed of PDSA, Point Local Aggregation Module (PLAM), and PDAM. In the first and second stages of PDNet, we implement PLAM by modifying the PointMetaBase block (defined in equation \ref{eq:5}) with additional ENE as follows:
\begin{equation}
    \begin{split}
    f_{i}^{l+1} = \mathcal{M}_{2}(\mathcal{A}(\{{{f^{l}_{j}}}^{'} + \delta(p_{j}^{l}-p_{i}^{l}) \\
    + \gamma(n_{j}^{l}), \forall j \in \mathcal{N}_i\})) + f_{i}^{l},
    \label{eq:13}
    \end{split}
\end{equation}
where $n_{j}^{l}$ is the point normals of point neighbor group of point $i$ at $l$-th stage, and $\gamma$ is the ENE implemented with MLP that map the input three dimensions of point normal vector to the dimension of high-level features. Incorporating with ENE, the PDAM defined in equation \ref{eq:11} can be further modified as:
\begin{equation}
    \begin{split}
            f_{i}^{l+1} = \mathcal{M}(\mathcal{A}(\{\Delta m_r^l f({{p^{l}_{r}}}^{'}) + \delta({{p^{l}_{r}}}^{'}-p_{i}^{l}) \\
            +\gamma(n_{r}^{l})\}_{r=1}^R )) + f_{i}^{l}.
    \label{eq:14}
    \end{split}
\end{equation}
We introduce parallel PLAM and PDAM in the third and fourth stages of PDNet. The point features are fed into PLAM to aggregate information locally (see equation \ref{eq:13}) and passed through PDAM (shown in equation \ref{eq:14}) to aggregate information globally at the same time. This design of MLP-based blocks with local and long-range dependencies helps our network learn strong generalization ability.

For a fair comparison, we adopt the same scaling strategies as \cite{lin2023meta, qian2022pointnext} to construct our PDNet. We define the number of deformable reference points $R=32$ to be consistent with the number of points in their local point groups. The configuration of three variants of PDNet is shown as follows:
\begin{itemize}
    \item PDNet-S: C = 32, B = 0
    \item PDNet-L: C = 32, B = (2, 4, 2, 2)
    \item PDNet-XXL: C = 64, B = (4, 8, 4, 4)
\end{itemize}
We denote C as the channel size of the stem MLP and B as the number of blocks in a stage. Notice that B = 0 means only one PDSA block but no PLAM or PDAM blocks are used at each stage.

\begin{table}[t]
\centering
\begin{tabular}{l|cc}
\hline
Method (time order) & mAcc (\%)  & OA (\%)\\
\hline
PointNet \cite{qi2017pointnet}  & 63.4 & 68.2 \\
PointNet++ \cite{qi2017pointnet++}  & 75.4 & 77.9 \\
PointCNN \cite{li2018pointcnn} & 75.1 & 78.5 \\
DGCNN \cite{wang2019dynamic} & 73.6 & 78.1 \\
PRA-Net \cite{cheng2021net} & 77.9 & 81.0 \\
PointMLP \cite{ma2022rethinking} & 84.4 & 85.7 \\
PointNeXt \cite{qian2022pointnext} & 86.8  & 88.2 \\
GAM \cite{hu2023gam} & 86.5  & 88.4\\
Point-PN \cite{zhang2023parameter} & - & 87.1 \\
PointMetaBase \cite{lin2023meta} & 86.8  & 88.2 \\
\hline
\textbf{PDNet} (ours) & \textbf{86.8} & \textbf{88.5} \\
\hline
\end{tabular}
\caption{Shape classification results on PB\_T50\_RS of ScanObjectNN. mAcc is the mean of class accuracy (\%) and OA is the overall accuracy (\%).}
\label{table1}
\end{table}

\section{Experiments}
In this section, we evaluate our PDNet on ScanObjectNN \cite{uy2019revisiting} for shape classification, S3DIS \cite{armeni20163d} for semantic segmentation, and ShapeNetPart \cite{yi2016scalable} for part segmentation. We also provide various ablation studies to better understand the PDNet.

\subsection{Classification and Segmentation}
\paragraph{Experimental setups.} We train our models by using CrossEntropy loss with label smoothing \cite{szegedy2016rethinking}, AdamW optimizer \cite{loshchilov2018decoupled}, an initial learning rate lr = 0.001, and weight decay $10^{-4}$ with Cosine Decay for all tasks. For S3DIS semantic segmentation task, point clouds are downsampled with a voxel size of 0.4 m following the previous methods \cite{zhao2021point, qian2021assanet, qian2022pointnext, lin2023meta}. For S3DIS, our PDNet is trained using a fixed number of 24000 points per batch with batch size set to 8 with an initial lr=0.01 for 100 epochs on a NVIDIA 3090 GPU and a 12-core Intel Xeon @ 2.50GHz CPU. For ScanObjectNN shape classification task, following \cite{qian2021assanet, lin2023meta}, our PDNet is trained by 1024 points with a weight decay of 0.05 for 250 epochs on a NVIDIA 3090 GPU. The points are randomly sampled during training and uniformly sampled during testing. For ShapeNetPart, we train our model using 2048 randomly sampled points with normals for 300 epochs on 4 NVIDIA 3090 GPUs. ShapeNetPart has the normal vectors of point clouds, so we do not apply the least square fitting to estimate the point normals. The original point normals are used for normal embedding. All the details of data augmentation are the same as those in PointNeXt \cite{qian2022pointnext} and PointMetaBase \cite{lin2023meta}.

\begin{table*}[t]
\centering
\begin{tabular}{l|ccc|ccc}
\hline
&\multicolumn{3}{c}{S3DIS 6-Fold} &\multicolumn{3}{c}{S3DIS Area-5}  \\
Method (time order)  & OA (\%) & mAcc (\%) & mIoU (\%) & OA (\%) & mAcc (\%) & mIoU (\%) \\
\hline
PointNet \cite{qi2017pointnet}  & 78.5 & 66.2 & 47.6 & - & 49.0 & 41.1\\
PointCNN \cite{li2018pointcnn} & 88.1 & 75.6 & 65.4 & 85.9 & 63.9 & 57.3\\
DGCNN \cite{wang2019dynamic} & 84.1 & - & 56.1 & 83.6 & - & 47.9\\
KPConv \cite{thomas2019kpconv} & - & 79.1 & 70.6 & - & 72.8 & 67.1\\
PCT \cite{guo2021pct} & - & 67.7 & 61.3 & - & - & -\\
PAConv \cite{xu2021paconv} & - & - & - & - & 73.0 & 66.6\\
AdaptConv \cite{zhou2021adaptive} & - & - & - & 90.0 & 73.2 & 67.9\\
Point Transformer \cite{zhao2021point} & 90.2 & 81.9 & 73.5 & 90.8 & 76.5 & 70.4\\
ASSANet \cite{qian2021assanet} & - & - & - & - & - & 66.8\\
CBL \cite{tang2022contrastive} & 89.6 & 79.4 & 73.1 & 90.6 & 75.2 & 69.4\\
StratifiedFormer \cite{lai2022stratified} & - & - & - & 91.5 & 78.1 & 72.0\\
Point TransformerV2 \cite{wu2022point} & - & - & - & 91.1 & 77.9 & 71.6\\
GAM \cite{hu2023gam} & 90.6 & 83.2 & 74.4 & - & - & -\\
\hline
PointNet++ \cite{qi2017pointnet++}  & 81.0 & 67.1 & 54.5 & 83.0 & - & 53.5\\
PointNeXt-L \cite{qian2022pointnext} & 89.8 & 82.2 & 73.9 & 90.0$\pm$0.1 & - & 69.0$\pm$0.5\\
PointNeXt-XL \cite{qian2022pointnext} & 90.3 & 83.0 & 74.9 & 90.6$\pm$0.1 & - & 70.5$\pm$0.3\\
PointMetaBase-L \cite{lin2023meta} & 90.6 & - & 75.6 & 90.5$\pm$0.1 & - & 69.5$\pm$0.3 \\
PointMetaBase-XXL \cite{lin2023meta} & 91.3 & - & 77.0 & 90.8$\pm$0.6 & - & 71.3$\pm$0.7 \\
\textbf{PDNet-L} (ours) & 91.4 & 85.5 & 76.7 & 90.7 & 77.1 & 70.8 \\
\textbf{PDNet-XXL} (ours) & \textbf{91.9} & \textbf{86.2} & \textbf{78.3} & \textbf{91.3} & \textbf{78.1}  & \textbf{72.3} \\
\hline
\end{tabular}
\caption{Semantic segmentation results on S3DIS (6-Fold and Area 5). OA is the overall accuracy (\%), mAcc is the mean of class accuracy (\%), and mIoU is the mean of instance IoU (\%).}
\label{table2}
\end{table*}

\begin{figure}[t]
\centering
\includegraphics[width=0.98\columnwidth]{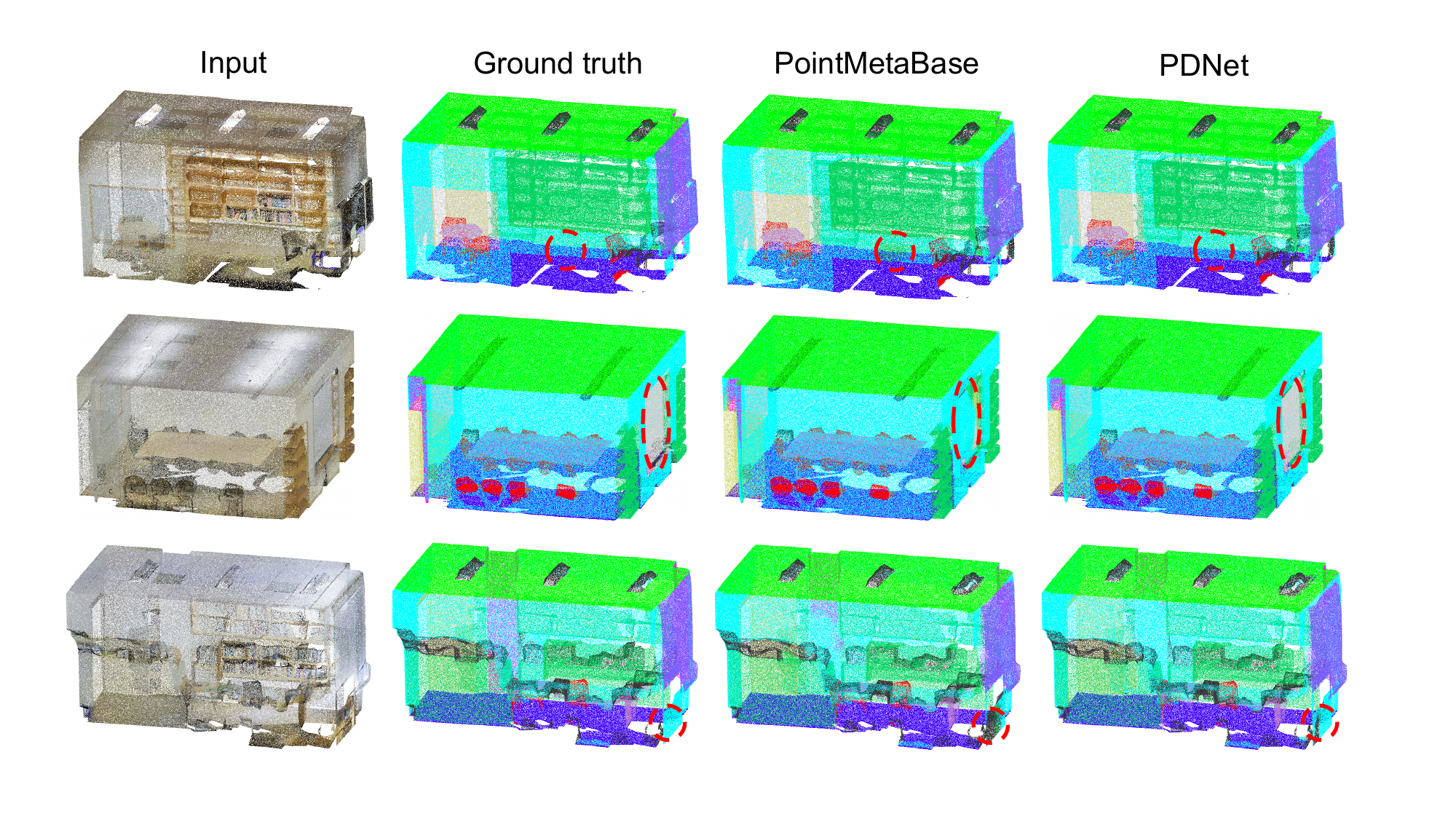} 
\caption{Visual comparison between MLP-based networks, PointMetaBase and our PDNet.}
\label{fig:S3DIS_visualization}
\end{figure}

\paragraph{Shape Classification.}  We first conduct experiments on a real-world shape classification dataset ScanobjectNN \cite{uy2019revisiting}. ScanObjectNN contains approximately 15,000 objects, which have 2902 unique instances that are categorized into 15 classes. We choose the hardest perturbed variant (PB\_T50\_RS) and report the overall accuracy (OA) and the mean of class accuracy (mAcc) results. As shown in Table \ref{table1}, our PDNet outperforms all baselines with the mAcc of 86.8\% and OA of 88.5\%. It shows point normals provide geometric information, and applying ENE helps improve the representation ability of the model.

\paragraph{Semantic Segmentation.} We also validate our PDNet on widely used Stanford Large-Scale 3D Indoor Spaces (S3DIS) \cite{armeni20163d} dataset for semantic segmentation task. S3DIS is a challenging benchmark that contains 271 rooms with 13 semantic categories in 6 areas. We report the OA, the mAcc, and the mean of instance IoU (mIoU) results of standard 6-fold cross-validation and Area-5 on S3DIS. As illustrated in Table \ref{table2}, Our PDNet-XXL outperforms all baselines with the OA of 91.9\%, mAcc of 86.2\%, and mIoU of 78.3\% on S3DIS 6-Fold and OA of 91.3\%, mAcc of 78.1\%, and mIoU of 72.3\% on S3DIS Area-5. Notably, the superior performance over powerful point Transformer architectures (StratifiedFormer and Point TransformerV2) shows the potential of MLP-based methods in point cloud analysis. Compared with recent MLP-based networks (PointNeXt and PointMetaBase), our PDNet-L gains +2.8\% and +1.1\% improvement in mIoU on S3DIS 6-Fold, respectively. Consistent progress is obtained when scaling up the models. It demonstrates the importance of long-range dependency in the point semantic segmentation task and the effectiveness of our method in aggregating information from deformable reference points conditioned on the input points. We also provide visualization of semantic segmentation results in Figure \ref{fig:S3DIS_visualization}, which clearly shows the superiority of our approach. Due to the direct modeling of long-range dependency, our method can recognize the objects in red circles while others fail.

\begin{figure}[t]
\centering
\includegraphics[width=0.96\columnwidth]{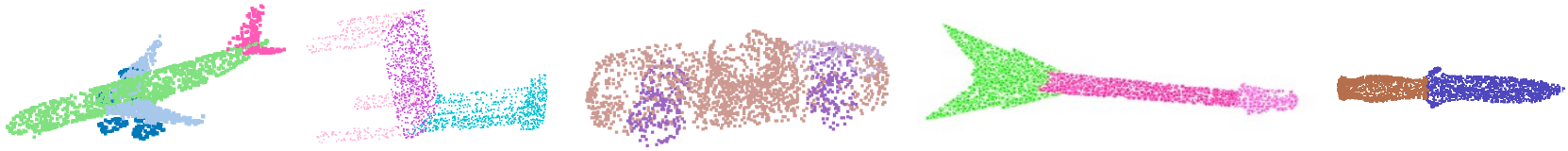} 
\caption{Visualization results on ShapeNetPart. }
\label{fig:shapenetpart_visual}
\end{figure}

\begin{table}[t]
\centering
\begin{tabular}{l|cc}
\hline
Method (time order) & c. mIoU  & i. mIoU  \\
\hline
PointNet \cite{qi2017pointnet}  & 80.4 & 83.7 \\
PointNet++ \cite{qi2017pointnet++}  & 81.9 & 85.1\\
SO-Net \cite{li2018so} & - & 84.6 \\
PointCNN \cite{li2018pointcnn} & 84.6 & 86.1 \\
DGCNN \cite{wang2019dynamic} & 82.3 & 85.1 \\
KPConv \cite{thomas2019kpconv} & 85.1 & 86.4 \\
PCT \cite{guo2021pct} & - & 86.4 \\
PAConv \cite{xu2021paconv} & 84.6 & 86.1 \\
AdaptConv \cite{zhou2021adaptive} & 83.4 & 86.4 \\
GDANet \cite{xu2021learning} & 85.0 & 86.5 \\
Point Trans. \cite{zhao2021point} & 83.7 & 86.6 \\
PointMLP \cite{ma2022rethinking}  & 84.6 & 86.1 \\
PointNeXt \cite{qian2022pointnext} & 85.2$\pm$0.1 & 87.0$\pm$0.1 \\
GAM \cite{hu2023gam} & - & 87.0 \\
Point-PN \cite{zhang2023parameter} & - & 86.6 \\
PointMetaBase \cite{lin2023meta} & 85.1$\pm$0.3 & 87.1$\pm$0.0 \\
\hline
\textbf{PDNet} (ours) & \textbf{85.4} & \textbf{87.2} \\
\hline
\end{tabular}
\caption{Part segmentation results on ShapeNetPart.}
\label{table3}
\end{table}

\paragraph{Part Segmentation.} ShapeNetPart \cite{yi2016scalable} is an object-level dataset for part segmentation. It contains 16,880 models with 16 different shape categories. Each category has 2-6 parts and up to 50 part labels in total. We evaluate the performance with the mean of class IoU (c. mIoU) and the mean of instance IoU (i. mIoU) in Table \ref{table3}. PDNet also achieves the best performance of 85.4\% in cls. mIoU and 87.2\% in mIoU. Visualization of part segmentation results are presented in Figure \ref{fig:shapenetpart_visual}.

\begin{table}[t]
\centering
\begin{tabular}{ccc|ccc}
\hline
PLAM & PDAM & NE  & mIoU & Params  & FLOPs\\
\hline
\checkmark & &  & 69.5$\pm$0.3  & 2.7 & 2.0\\
 & \checkmark&   & 70.1$\pm$0.2 &3.6 &2.1\\
 \checkmark & &\checkmark   & 69.7$\pm$0.3 &2.7 &2.0 \\
 \checkmark &\checkmark &   & 70.4$\pm$0.3 &4.9 &2.4\\
 \checkmark &\checkmark &\checkmark   & \textbf{70.6$\pm$0.2} &4.9 &2.4\\
\hline
\end{tabular}
\caption{Evaluation of proposed components on S3DIS Area-5. mIoU is the mean of IoU (\%).}
\label{table4}
\end{table}

\subsection{Ablation Studies}
\paragraph{Effectiveness of Proposed Components.} We evaluate the performance of proposed components of PDNet-L in Table \ref{table4} with mean±std in three random runs. PLAM and PDAM in the first and second column of Table \ref{table4} represents whether to use them in the third and fourth stages of PDNet-L. The results show that using PDAM to aggregate information from the deformable reference regions is better than adopting PLAM to aggregate information within local point groups. It demonstrates that the network prefers to learn global relations in the deep stages. In the case of using both PLAM and PDAM, it obtains better result than only use PLAM or PDAM. Our PDNet-L with all the proposed components achieves the best performance of 70.6±0.2\% in mIoU.

\begin{table}[t]
\centering
\begin{tabular}{ccc|c}
\hline
Stage2 & Stage3 & Stage4  & mIoU (\%) \\
\hline
 & &  & 69.7 \\
  & & \checkmark & 70.1 \\
   &\checkmark & \checkmark & \textbf{70.8}\\
\checkmark  &\checkmark & \checkmark & 69.9\\
\hline
\end{tabular}
\caption{Ablation study on applying PDAM in different stages on S3DIS Area-5.}
\label{table5}
\end{table}

\begin{table}[t]
\centering
\begin{tabular}{c|c}
\hline
method & mIoU (\%) \\
\hline
random & 70.1\\
center & 70.6\\
FPS & \textbf{70.8}\\
\hline
 successive & 70.4 \\
parallel &   \textbf{70.8}\\
\hline
\end{tabular}
\caption{Ablation study on different types of initial reference points and ablation study on combining strategy of PDAM and PLAM on S3DIS Area-5.}
\label{table6}
\end{table}

\paragraph{PDAM.} We first explore adopting Point Deformable Aggregation Module (PDAM) at different stages. As shown in Table \ref{table5}, only adopting PDAM in the last stage improves by 0.4\% and applying it in the last two stages leads to the best performance of 70.8\% in mIoU. However, using PDAM at the early stage obtains decreasement in mIoU, which reveals the cues that our network performs better when adopting Point Local Aggregation Module (PLAM) in the early stages to capture local geometries and PDAM in the deeper stages to model long-range dependencies.

We also investigate several types of initializing methods for the reference points on S3DIS Area-5. The results are presented in Table \ref{table6}, which suggests that using FPS to acquire the initial reference point based on the input point positions is superior to random initialization. Moreover, using the center of the input points as the initial status for all the initial reference points performs worse than considering each reference point independently to be uniformly distributed in space. We further conduct an ablation study on combining strategy of PDAM and PLAM in Table \ref{table6}. For 3D point clouds, applying PLAM and PDAM to aggregate information from local and distant regions at the same time is better than widely used successively designed \cite{chu2021twins, yang2022moat, xia2022vision} in 2D images that adopt PDAM to model long-range relations after capturing local features by PLAM.

\section{Conclusion}
In this paper,  we propose PDNet, a concise MLP-based network for point cloud processing. Equipped with Point Deformable Aggregation Module (PDAM), our model achieves both long-range dependency and adaptive spatial aggregation in a data-dependent way. For each query point, PDAM aggregates information from deformable reference points, which are initialized according to the point positions and then shifted via additional offsets and modulation scalars conditioned on the input point features. Enhanced Normal Embedding further helps improve the representation ability of point itself. Extensive experiments and ablation studies illustrate the effectiveness of PDNet over various tasks. We hope our work can inspire insights toward exploring suitable deformable mechanisms for point clouds.


\bibliography{aaai24}

\begin{thebibliography}{46}
\providecommand{\natexlab}[1]{#1}

\bibitem[{Armeni et~al.(2016)Armeni, Sener, Zamir, Jiang, Brilakis, Fischer, and Savarese}]{armeni20163d}
Armeni, I.; Sener, O.; Zamir, A.~R.; Jiang, H.; Brilakis, I.; Fischer, M.; and Savarese, S. 2016.
\newblock 3d semantic parsing of large-scale indoor spaces.
\newblock In \emph{Proceedings of the IEEE Conference on Computer Vision and Pattern Recognition}, 1534--1543.

\bibitem[{Cheng et~al.(2021)Cheng, Chen, He, Liu, and Bai}]{cheng2021net}
Cheng, S.; Chen, X.; He, X.; Liu, Z.; and Bai, X. 2021.
\newblock Pra-net: Point relation-aware network for 3d point cloud analysis.
\newblock \emph{IEEE Transactions on Image Processing}, 30: 4436--4448.

\bibitem[{Chu et~al.(2021)Chu, Tian, Wang, Zhang, Ren, Wei, Xia, and Shen}]{chu2021twins}
Chu, X.; Tian, Z.; Wang, Y.; Zhang, B.; Ren, H.; Wei, X.; Xia, H.; and Shen, C. 2021.
\newblock Twins: Revisiting the design of spatial attention in vision transformers.
\newblock \emph{Advances in Neural Information Processing Systems}, 34: 9355--9366.

\bibitem[{Dai et~al.(2017)Dai, Qi, Xiong, Li, Zhang, Hu, and Wei}]{dai2017deformable}
Dai, J.; Qi, H.; Xiong, Y.; Li, Y.; Zhang, G.; Hu, H.; and Wei, Y. 2017.
\newblock Deformable convolutional networks.
\newblock In \emph{Proceedings of the IEEE International Conference on Computer Vision}, 764--773.

\bibitem[{Goyal et~al.(2021)Goyal, Law, Liu, Newell, and Deng}]{goyal2021revisiting}
Goyal, A.; Law, H.; Liu, B.; Newell, A.; and Deng, J. 2021.
\newblock Revisiting point cloud shape classification with a simple and effective baseline.
\newblock In \emph{International Conference on Machine Learning}, 3809--3820. PMLR.

\bibitem[{Guo et~al.(2021)Guo, Cai, Liu, Mu, Martin, and Hu}]{guo2021pct}
Guo, M.-H.; Cai, J.-X.; Liu, Z.-N.; Mu, T.-J.; Martin, R.~R.; and Hu, S.-M. 2021.
\newblock Pct: Point cloud transformer.
\newblock \emph{Computational Visual Media}, 7: 187--199.

\bibitem[{Hu et~al.(2023)Hu, Fanyi, Jingwen, Hongtao, Yaonong, Laifeng, Yanhao, and Zhiwang}]{hu2023gam}
Hu, H.; Fanyi, W.; Jingwen, S.; Hongtao, Z.; Yaonong, W.; Laifeng, H.; Yanhao, Z.; and Zhiwang, Z. 2023.
\newblock GAM : Gradient Attention Module of Optimization for Point Clouds Analysis.
\newblock In \emph{Proceedings of the AAAI Conference on Artificial Intelligence}, volume~37, 835--843.

\bibitem[{Lai et~al.(2022)Lai, Liu, Jiang, Wang, Zhao, Liu, Qi, and Jia}]{lai2022stratified}
Lai, X.; Liu, J.; Jiang, L.; Wang, L.; Zhao, H.; Liu, S.; Qi, X.; and Jia, J. 2022.
\newblock Stratified transformer for 3d point cloud segmentation.
\newblock In \emph{Proceedings of the IEEE/CVF Conference on Computer Vision and Pattern Recognition}, 8500--8509.

\bibitem[{Li, Chen, and Lee(2018)}]{li2018so}
Li, J.; Chen, B.~M.; and Lee, G.~H. 2018.
\newblock So-net: Self-organizing network for point cloud analysis.
\newblock In \emph{Proceedings of the IEEE Conference on Computer Vision and Pattern Recognition}, 9397--9406.

\bibitem[{Li et~al.(2018)Li, Bu, Sun, Wu, Di, and Chen}]{li2018pointcnn}
Li, Y.; Bu, R.; Sun, M.; Wu, W.; Di, X.; and Chen, B. 2018.
\newblock Pointcnn: Convolution on x-transformed points.
\newblock \emph{Advances in Neural Information Processing Systems}, 31.

\bibitem[{Lin et~al.(2023)Lin, Zheng, Li, Chao, Wang, Wang, Tian, and Ji}]{lin2023meta}
Lin, H.; Zheng, X.; Li, L.; Chao, F.; Wang, S.; Wang, Y.; Tian, Y.; and Ji, R. 2023.
\newblock Meta Architecture for Point Cloud Analysis.
\newblock In \emph{Proceedings of the IEEE/CVF Conference on Computer Vision and Pattern Recognition}, 17682--17691.

\bibitem[{Liu et~al.(2019)Liu, Fan, Xiang, and Pan}]{liu2019relation}
Liu, Y.; Fan, B.; Xiang, S.; and Pan, C. 2019.
\newblock Relation-shape convolutional neural network for point cloud analysis.
\newblock In \emph{Proceedings of the IEEE/CVF Conference on Computer Vision and Pattern Recognition}, 8895--8904.

\bibitem[{Loshchilov and Hutter(2018)}]{loshchilov2018decoupled}
Loshchilov, I.; and Hutter, F. 2018.
\newblock Decoupled Weight Decay Regularization.
\newblock In \emph{International Conference on Learning Representations}.

\bibitem[{Ma et~al.(2022)Ma, Qin, You, Ran, and Fu}]{ma2022rethinking}
Ma, X.; Qin, C.; You, H.; Ran, H.; and Fu, Y. 2022.
\newblock Rethinking network design and local geometry in point cloud: A simple residual MLP framework.
\newblock \emph{arXiv preprint arXiv:2202.07123}.

\bibitem[{Maturana and Scherer(2015)}]{maturana2015voxnet}
Maturana, D.; and Scherer, S. 2015.
\newblock Voxnet: A 3d convolutional neural network for real-time object recognition.
\newblock In \emph{2015 IEEE/RSJ International Conference on Intelligent Robots and Systems (IROS)}, 922--928. IEEE.

\bibitem[{Mitra and Nguyen(2003)}]{mitra2003estimating}
Mitra, N.~J.; and Nguyen, A. 2003.
\newblock Estimating surface normals in noisy point cloud data.
\newblock In \emph{Proceedings of the Nineteenth Annual Symposium on Computational Geometry}, 322--328.

\bibitem[{Qi et~al.(2017{\natexlab{a}})Qi, Su, Mo, and Guibas}]{qi2017pointnet}
Qi, C.~R.; Su, H.; Mo, K.; and Guibas, L.~J. 2017{\natexlab{a}}.
\newblock Pointnet: Deep learning on point sets for 3d classification and segmentation.
\newblock In \emph{Proceedings of the IEEE Conference on Computer Vision and Pattern Recognition}, 652--660.

\bibitem[{Qi et~al.(2017{\natexlab{b}})Qi, Yi, Su, and Guibas}]{qi2017pointnet++}
Qi, C.~R.; Yi, L.; Su, H.; and Guibas, L.~J. 2017{\natexlab{b}}.
\newblock Pointnet++: Deep hierarchical feature learning on point sets in a metric space.
\newblock \emph{Advances in Neural Information Processing Systems}, 30.

\bibitem[{Qian et~al.(2021)Qian, Hammoud, Li, Thabet, and Ghanem}]{qian2021assanet}
Qian, G.; Hammoud, H.; Li, G.; Thabet, A.; and Ghanem, B. 2021.
\newblock Assanet: An anisotropic separable set abstraction for efficient point cloud representation learning.
\newblock \emph{Advances in Neural Information Processing Systems}, 34: 28119--28130.

\bibitem[{Qian et~al.(2022)Qian, Li, Peng, Mai, Hammoud, Elhoseiny, and Ghanem}]{qian2022pointnext}
Qian, G.; Li, Y.; Peng, H.; Mai, J.; Hammoud, H.; Elhoseiny, M.; and Ghanem, B. 2022.
\newblock Pointnext: Revisiting pointnet++ with improved training and scaling strategies.
\newblock \emph{Advances in Neural Information Processing Systems}, 35: 23192--23204.

\bibitem[{Ran, Liu, and Wang(2022)}]{ran2022surface}
Ran, H.; Liu, J.; and Wang, C. 2022.
\newblock Surface representation for point clouds.
\newblock In \emph{Proceedings of the IEEE/CVF Conference on Computer Vision and Pattern Recognition}, 18942--18952.

\bibitem[{Su et~al.(2015)Su, Maji, Kalogerakis, and Learned-Miller}]{su2015multi}
Su, H.; Maji, S.; Kalogerakis, E.; and Learned-Miller, E. 2015.
\newblock Multi-view convolutional neural networks for 3d shape recognition.
\newblock In \emph{Proceedings of the IEEE International Conference on Computer Vision}, 945--953.

\bibitem[{Szegedy et~al.(2016)Szegedy, Vanhoucke, Ioffe, Shlens, and Wojna}]{szegedy2016rethinking}
Szegedy, C.; Vanhoucke, V.; Ioffe, S.; Shlens, J.; and Wojna, Z. 2016.
\newblock Rethinking the inception architecture for computer vision.
\newblock In \emph{Proceedings of the IEEE Conference on Computer Vision and Pattern Recognition}, 2818--2826.

\bibitem[{Tang et~al.(2022{\natexlab{a}})Tang, Zhan, Chen, Yu, and Tao}]{tang2022contrastive}
Tang, L.; Zhan, Y.; Chen, Z.; Yu, B.; and Tao, D. 2022{\natexlab{a}}.
\newblock Contrastive boundary learning for point cloud segmentation.
\newblock In \emph{Proceedings of the IEEE/CVF Conference on Computer Vision and Pattern Recognition}, 8489--8499.

\bibitem[{Tang et~al.(2022{\natexlab{b}})Tang, Qian, Zhang, Zeng, Hou, and Zhe}]{tang2022warpinggan}
Tang, Y.; Qian, Y.; Zhang, Q.; Zeng, Y.; Hou, J.; and Zhe, X. 2022{\natexlab{b}}.
\newblock WarpingGAN: Warping multiple uniform priors for adversarial 3D point cloud generation.
\newblock In \emph{Proceedings of the IEEE/CVF Conference on Computer Vision and Pattern Recognition}, 6397--6405.

\bibitem[{Thomas et~al.(2019)Thomas, Qi, Deschaud, Marcotegui, Goulette, and Guibas}]{thomas2019kpconv}
Thomas, H.; Qi, C.~R.; Deschaud, J.-E.; Marcotegui, B.; Goulette, F.; and Guibas, L.~J. 2019.
\newblock Kpconv: Flexible and deformable convolution for point clouds.
\newblock In \emph{Proceedings of the IEEE/CVF International Conference on Computer Vision}, 6411--6420.

\bibitem[{Uy et~al.(2019)Uy, Pham, Hua, Nguyen, and Yeung}]{uy2019revisiting}
Uy, M.~A.; Pham, Q.-H.; Hua, B.-S.; Nguyen, T.; and Yeung, S.-K. 2019.
\newblock Revisiting point cloud classification: A new benchmark dataset and classification model on real-world data.
\newblock In \emph{Proceedings of the IEEE/CVF International Conference on Computer Vision}, 1588--1597.

\bibitem[{Wang, Samari, and Siddiqi(2018)}]{wang2018local}
Wang, C.; Samari, B.; and Siddiqi, K. 2018.
\newblock Local spectral graph convolution for point set feature learning.
\newblock In \emph{Proceedings of the European Conference on Computer Vision (ECCV)}, 52--66.

\bibitem[{Wang et~al.(2023)Wang, Dai, Chen, Huang, Li, Zhu, Hu, Lu, Lu, Li et~al.}]{wang2023internimage}
Wang, W.; Dai, J.; Chen, Z.; Huang, Z.; Li, Z.; Zhu, X.; Hu, X.; Lu, T.; Lu, L.; Li, H.; et~al. 2023.
\newblock Internimage: Exploring large-scale vision foundation models with deformable convolutions.
\newblock In \emph{Proceedings of the IEEE/CVF Conference on Computer Vision and Pattern Recognition}, 14408--14419.

\bibitem[{Wang et~al.(2018)Wang, Girshick, Gupta, and He}]{wang2018non}
Wang, X.; Girshick, R.; Gupta, A.; and He, K. 2018.
\newblock Non-local neural networks.
\newblock In \emph{Proceedings of the IEEE Conference on Computer Vision and Pattern Recognition}, 7794--7803.

\bibitem[{Wang et~al.(2019)Wang, Sun, Liu, Sarma, Bronstein, and Solomon}]{wang2019dynamic}
Wang, Y.; Sun, Y.; Liu, Z.; Sarma, S.~E.; Bronstein, M.~M.; and Solomon, J.~M. 2019.
\newblock Dynamic graph cnn for learning on point clouds.
\newblock \emph{ACM Transactions on Graphics (ToG)}, 38(5): 1--12.

\bibitem[{Wu, Qi, and Fuxin(2019)}]{wu2019pointconv}
Wu, W.; Qi, Z.; and Fuxin, L. 2019.
\newblock Pointconv: Deep convolutional networks on 3d point clouds.
\newblock In \emph{Proceedings of the IEEE/CVF Conference on Computer Vision and Pattern Recognition}, 9621--9630.

\bibitem[{Wu et~al.(2022)Wu, Lao, Jiang, Liu, and Zhao}]{wu2022point}
Wu, X.; Lao, Y.; Jiang, L.; Liu, X.; and Zhao, H. 2022.
\newblock Point transformer v2: Grouped vector attention and partition-based pooling.
\newblock \emph{Advances in Neural Information Processing Systems}, 35: 33330--33342.

\bibitem[{Wu et~al.(2015)Wu, Song, Khosla, Yu, Zhang, Tang, and Xiao}]{wu20153d}
Wu, Z.; Song, S.; Khosla, A.; Yu, F.; Zhang, L.; Tang, X.; and Xiao, J. 2015.
\newblock 3d shapenets: A deep representation for volumetric shapes.
\newblock In \emph{Proceedings of the IEEE Conference on Computer Vision and Pattern Recognition}, 1912--1920.

\bibitem[{Xia et~al.(2022)Xia, Pan, Song, Li, and Huang}]{xia2022vision}
Xia, Z.; Pan, X.; Song, S.; Li, L.~E.; and Huang, G. 2022.
\newblock Vision transformer with deformable attention.
\newblock In \emph{Proceedings of the IEEE/CVF Conference on Computer Vision and Pattern Recognition}, 4794--4803.

\bibitem[{Xu et~al.(2021{\natexlab{a}})Xu, Ding, Zhao, and Qi}]{xu2021paconv}
Xu, M.; Ding, R.; Zhao, H.; and Qi, X. 2021{\natexlab{a}}.
\newblock Paconv: Position adaptive convolution with dynamic kernel assembling on point clouds.
\newblock In \emph{Proceedings of the IEEE/CVF Conference on Computer Vision and Pattern Recognition}, 3173--3182.

\bibitem[{Xu et~al.(2021{\natexlab{b}})Xu, Zhang, Zhou, Xu, Qi, and Qiao}]{xu2021learning}
Xu, M.; Zhang, J.; Zhou, Z.; Xu, M.; Qi, X.; and Qiao, Y. 2021{\natexlab{b}}.
\newblock Learning geometry-disentangled representation for complementary understanding of 3d object point cloud.
\newblock In \emph{Proceedings of the AAAI Conference on Artificial Intelligence}, volume~35, 3056--3064.

\bibitem[{Yang et~al.(2022)Yang, Qiao, Yu, Yuan, Zhu, Yuille, Adam, and Chen}]{yang2022moat}
Yang, C.; Qiao, S.; Yu, Q.; Yuan, X.; Zhu, Y.; Yuille, A.; Adam, H.; and Chen, L.-C. 2022.
\newblock MOAT: Alternating Mobile Convolution and Attention Brings Strong Vision Models.
\newblock In \emph{International Conference on Learning Representations}.

\bibitem[{Yang et~al.(2020)Yang, Sun, Liu, Qi, and Jia}]{yang2020cn}
Yang, Z.; Sun, Y.; Liu, S.; Qi, X.; and Jia, J. 2020.
\newblock Cn: Channel normalization for point cloud recognition.
\newblock In \emph{Proceedings of the European Conference on Computer Vision (ECCV)}, 600--616.

\bibitem[{Yi et~al.(2016)Yi, Kim, Ceylan, Shen, Yan, Su, Lu, Huang, Sheffer, and Guibas}]{yi2016scalable}
Yi, L.; Kim, V.~G.; Ceylan, D.; Shen, I.-C.; Yan, M.; Su, H.; Lu, C.; Huang, Q.; Sheffer, A.; and Guibas, L. 2016.
\newblock A scalable active framework for region annotation in 3d shape collections.
\newblock \emph{ACM Transactions on Graphics (ToG)}, 35(6): 1--12.

\bibitem[{Yue et~al.(2021)Yue, Sun, Kuang, Wei, Torr, Zhang, and Lin}]{yue2021vision}
Yue, X.; Sun, S.; Kuang, Z.; Wei, M.; Torr, P.~H.; Zhang, W.; and Lin, D. 2021.
\newblock Vision transformer with progressive sampling.
\newblock In \emph{Proceedings of the IEEE/CVF International Conference on Computer Vision}, 387--396.

\bibitem[{Zhang et~al.(2023)Zhang, Wang, Wang, Gao, Li, and Shi}]{zhang2023parameter}
Zhang, R.; Wang, L.; Wang, Y.; Gao, P.; Li, H.; and Shi, J. 2023.
\newblock Parameter is not all you need: Starting from non-parametric networks for 3d point cloud analysis.
\newblock arXiv:2303.08134.

\bibitem[{Zhao et~al.(2021)Zhao, Jiang, Jia, Torr, and Koltun}]{zhao2021point}
Zhao, H.; Jiang, L.; Jia, J.; Torr, P.~H.; and Koltun, V. 2021.
\newblock Point transformer.
\newblock In \emph{Proceedings of the IEEE/CVF International Conference on Computer Vision}, 16259--16268.

\bibitem[{Zhou et~al.(2021)Zhou, Feng, Fang, Wei, Qin, and Lu}]{zhou2021adaptive}
Zhou, H.; Feng, Y.; Fang, M.; Wei, M.; Qin, J.; and Lu, T. 2021.
\newblock Adaptive graph convolution for point cloud analysis.
\newblock In \emph{Proceedings of the IEEE/CVF International Conference on Computer Vision}, 4965--4974.

\bibitem[{Zhu et~al.(2019)Zhu, Hu, Lin, and Dai}]{zhu2019deformable}
Zhu, X.; Hu, H.; Lin, S.; and Dai, J. 2019.
\newblock Deformable convnets v2: More deformable, better results.
\newblock In \emph{Proceedings of the IEEE/CVF Conference on Computer Vision and Pattern Recognition}, 9308--9316.

\bibitem[{Zhu et~al.(2020)Zhu, Su, Lu, Li, Wang, and Dai}]{zhu2020deformable}
Zhu, X.; Su, W.; Lu, L.; Li, B.; Wang, X.; and Dai, J. 2020.
\newblock Deformable DETR: Deformable Transformers for End-to-End Object Detection.
\newblock In \emph{International Conference on Learning Representations}.

\end{thebibliography}

\end{document}